\documentclass[preprint,12pt]{article}
\usepackage{epsfig}
\usepackage{amsmath}
\usepackage{amssymb}
\usepackage{caption} 
\usepackage{epstopdf}
\usepackage{algorithm}
\usepackage{algorithmic}
\usepackage{placeins} % for FloatBarrier
\usepackage{graphicx}
\usepackage{multicol,array}
\usepackage{multirow}
\usepackage[T1]{fontenc}
\usepackage{lmodern}
\usepackage[top=1.0in, bottom=1.0in, left=1.0in, right=1.0in]{geometry}
\begin{document}
	{\centering
		
		{\bfseries\Large A self-tuning Firefly algorithm to tune the parameters of Ant Colony System (ACSFA)\bigskip}
		
	M. K. A. Ariyaratne\textsuperscript{1} , T. G. I. Fernando\textsuperscript{2} and S. Weerakoon\textsuperscript{3}  \\
	{\itshape
			\textsuperscript{1}Department of Computer Science, Faculty of Computing, General Sir John Kotelawala Defence University, Sri Lanka.\\
			\textsuperscript{2}Department of Computer Science, Faculty of Applied Sciences, University of Sri Jayewardenepura, Sri Lanka. \\
			\textsuperscript{3} Department of Mathematics, Faculty of Applied Sciences, University of Sri Jayewardenepura, Sri Lanka. \\

	}}
\begin{abstract}
Ant colony system (ACS) is a promising approach which has been widely used in problems such as Travelling Salesman Problems (TSP), Job shop scheduling problems (JSP) and Quadratic Assignment problems (QAP). In its original implementation, parameters of the algorithm were selected by trial and error approach. Over the last few years,  novel approaches have been proposed on adapting the parameters of ACS in improving its performance. The aim of this paper is to use a framework introduced for self-tuning optimization algorithms combined with the firefly algorithm (FA) to tune the parameters of the ACS solving symmetric TSP problems. The FA optimizes the problem specific parameters of ACS while the parameters of the FA are tuned by the selected framework itself.  With this approach, the user neither has to work with the parameters of ACS nor the parameters of FA. Using common symmetric TSP problems we demonstrate that the framework fits well for the ACS. A detailed statistical analysis further verifies the goodness of the new ACS over the existing ACS and also of the other techniques used to tune the parameters of ACS.

\end{abstract}

%\begin{keyword}
%Self tuning framework \sep Ant Colony System \sep Travelling Salesman problem \sep  Firefly Algorithm 

\section{Introduction}
The collective behavior of natural insects - ants, bees, fireflies and termites mimic the problem solving capabilities of the swarms \cite{Deborah, Morse, DeCock2005807}. These capabilities were adopted in various heuristics and meta-heuristics to solve difficult optimization problems. Each meta-heuristic has a real world inspiration of optimization. As such, the main inspiration for the ant colony system algorithm is the natural food finding strategy of ants. Ants are capable of finding the shortest path from the food source to their nests. A chemical inside them call pheromone is the reason for this optimized behavior. Following this real world strategy, the ant colony system algorithm was developed by Marco Dorigo et al.\cite{Dorigo:1997:ACS:2221336.2221407} to suit for path optimization problems. Initially the algorithm was developed to solve the TSP. This original implementation supports the hypothesis that the ant colony algorithm is successful in finding the shortest path for the TSP.

Similar to any meta-heuristic, the ant colony system also has algorithm specific parameters. The initial implementation used the trial and error method to find the best collection of parameters. The values of the parameters depend on the problem at hand and hence at each instance, the most suitable parameter set for the problem should be evaluated. The task of parameter tuning again is an optimization where the optimal parameter set will give the optimum performance.

Though the initial study relied on trial and error, Dorigo had stated some important features of parameters such as pheromone behavior, the number of ants and how they affect the performance of the algorithm \cite{Dorigo:1997:ACS:2221336.2221407}. The original paper discusses the ranges for different parameters so that an idea about the distribution of each parameter is given. The results were in an encouraging position ascertaining that the algorithm is successful in solving the TSP. Instances from TSPLIB and randomly generated TSPs' were used to evaluate the algorithm.

Since the original ACS works well with pre-tested parameter values, finding approaches to set better parameters without trial and error can improve the performance of the algorithm. The best way is to consider setting of parameters as another optimization problem. The same matter of tuning parameters of an ACS is considered by many other researchers. Yet none was successful in maintaining fully parameterless environment. An Adaptive Parameter Control Strategy for Ant Colony Systems by Zhi-Feng Hao et al. is a study carried out to enhance the performance of the ant system solving the TSP \cite{4028059}. Although it has been mentioned as the ant colony system, they have used the ant system for the study. The particle swarm optimization (PSO) has been used to optimize the parameters of the ant system (PSOACS). The parameters of the ant system were considered as a whole where one particle represents approximation for the set of parameters. The number of particles is the same as the number of ants, and once ants complete a tour the PSO will update the parameters. The conclusions are based on the performance of the new algorithm and the existing ACS. The results support the conclusions, but the following matters were not addressed. Although the study focused on a parameter tuning technique, the effects of the method towards the parameters were not mentioned. Basically the study was focused on improving the TSP results of the existing approach. For both PSOACS and original ACS, the results obtained conflict with the optimal results given in the TSPLIB. The reason may be the differences in the implementations. Also in PSOACS, the equations used are not clear enough to get an idea about the new algorithm.

In Evolving Ant Colony Optimization, another research by Hozefa M. Botee and Eric Bonabeauy, they have made the use of genetic algorithm (GA) to evolve the best set of parameters \cite{doi:10.1142/S0219525998000119}. Here also, one parameter set represents an individual of the genetic algorithm. However the implementation was tested only over two TSP instances. The results were encouraging, but the parameters of the ACS depend on the selection of the parameters of GA such as crossover and mutation probability.

Apart from meta-heuristics, machine learning techniques have also been used to tune parameters of ACS. For example Ayse Hande Erol et al. in their research, have used artificial neural networks (ANN) to find the best parameter set for ACS solving a given TSP \cite{erol2012optimizing}. The research focuses on only two parameters, $\alpha$: pheromone decay parameter and $\beta$: the relative importance of the pheromone vs. distance. Initially the ACS-ANN hybrid algorithm runs with different $\alpha$ and $\beta$ values for 50 times. These parameters work as the inputs to the ANN. Then ANN predicts the best parameter values for a given TSP instance. The hybrid algorithm was tested with several TSP instances from the TSPLIB \cite{tsplib}. The results support the hypothesis that the hybrid algorithm performs well in finding better parameter values. However the study focused on tuning only two parameters. The information about the ANN such as the training methods, weights and their effects are not mentioned in the research.

As such there are other researches which have been conducted to find better parameter sets for the ACS in solving TSP \cite{gomez2005fine, zitar2005optimizing}. But as a whole, all these methods rely on another algorithm or technique, where it again contains its own parameters. Therefore the performance of the ACS again depends on the parameters of the used algorithm. To overcome this issue we have designed an algorithm with the help of the firefly algorithm and the self-tuning framework for optimization algorithms proposed by Xin-She Yang, to optimize the parameters of the ACS algorithm solving symmetric TSP instances. Xin-She Yang et al., in the implementation of the self-tuning framework have used the firefly algorithm to apply the framework to tune FA's parameters \cite{yang2013framework}. In their framework, the problem solved by the optimization algorithm and the parameter set of the algorithm both were considered as a single problem. The framework was initially tested for the firefly algorithm and proved its capability of tuning parameters of itself (FA). In a research done by M.K.A. Ariyaratne et al., use this self tuning framework combined with the firefly algorithm to solve nonlinear equitations \cite{ariyaratne}. They have solved univariate nonlinear equations having complex roots with the help of a modified firefly algorithm. To find the best  parameters values, the self tuning framework was implemented on the firefly algorithm. In their research, a firefly carries an approximation for a root as well as approximations to the parameter values. Finally as the output, they receive best approximations for the roots in a given range as well as best approximations for the parameter values. The research again confirms the powerfulness of the self tuning framework in optimizing parameters. 

The significance of this research lies in the potential of the developed ant colony
optimization algorithm for the TSP with the self tuning framework to optimize the parameters. Despite the recent advancements in the field of route optimization and parameter optimization, the following issues have not been addressed by other researchers (See Table \ref{tab: tab1} for details of previously applied approaches) where our research has accomplished.\\
\begin{itemize}
	\item None of the existing systems are capable of providing virtually parameter-free environments for the ant colony systems to solve TSPs.
	\item In most of the researches, parameter optimization is done using another meta-heuristic algorithm whose parameters should be manually selected.
	\item None of the approaches have used a designed framework for parameter optimization.
\end{itemize}
\FloatBarrier
		%\hspace*{-1.5cm}
	\begin{table}[h]
		
		\renewcommand{\arraystretch}{1.4}
		
		\centering
		\hskip-2.0cm
		\footnotesize 
		\begin{center}	
		\begin{tabular}{|p{3.8cm}|p{0.8cm}|p{2.65cm}|p{1cm}|p{3.3cm}|p{4cm}|}
			
			\hline
			\textbf{Author}                                                                 & \textbf{Year} & \textbf{Parameter Optimization method} & \textbf{\# of TSP problems tested} & \textbf{Advantages}                                                           & \textbf{Limitations}                                                                                                            \\ \hline
			Hozefa M. Botee and Eric Bonabeauy    & 1998          & GA                                                                               & 2   & GA optimizes the parameters of ACS                                           & \parbox{3.6cm}{Only some parameters of ACS were optimized.\\The parameters of GA should be manually updated.}
			\\ \hline
			Raed Abu Zaitar and Hussein Hiyassat  & 2005          & GA                                                                               & 8                                                                                & GA optimizes the parameters of ACS & Only two parameters of ACS ($\alpha$ and $\beta$) were updated in the first phase.                                                      \\ \hline
			D. Gómez-Cabrero and D. N. Ranasinghe & 2005          & PSO                                                                              & 24                                                                               & PSO optimizes the parameters of ACS & \parbox{3.6cm}{The parameters of PSO should be manually updated.\\High computational overhead. }        \\ \hline
			Zhi-Feng Hao et al.                                                             & 2006          & PSO                                                                              & 10                                                                               & \begin{tabular}[c]{@{}l@{}}PSO optimizes the\\ parameters of ACS\end{tabular} & The parameters of PSO should be manually updated.                                                                                     \\ \hline
			Ayse Hande Erol et al.                                                          & 2012          & \begin{tabular}[c]{@{}l@{}}Artificial Neural\\ Networks (ANN)\end{tabular}       & 16                                                                               & \begin{tabular}[c]{@{}l@{}}ANN optimizes the\\ parameters of ACO\end{tabular} & Only some parameters of ACO were optimized.                                                                                     \\ \hline
		\end{tabular}

		\caption{Previous work on parameter optimization of Ant Systems\\~\\ GA-Genetic Algorithms, PSO-Particle Swarm Optimization, ACS-Ant Colony Systems, ACO-Ant Colony Optimization}
			\end{center}
		\label{tab: tab1}
	\end{table}

\FloatBarrier
In this paper, Yang's self-tuning framework is studied, and a parameter selection strategy based on the firefly algorithm is developed. To deliver a better idea about the present work, the remainder of this paper is structured as follows. Section 2 provides some preliminaries related to the Ant colony system and the firefly algorithm. In section 3, we briefly describe the self-tuning framework for the firefly algorithm and how we adopt it to tune the parameters of the ACS when solving symmetric TSPs. There, we also present the hybrid ACS-FA. Section 4 points out the TSP examples, the parameters used and the results obtained from the new approach. To emphasize the goodness of the new algorithm, we compare the results with the original ACS and with some other parameter tuning approaches. Finally, we draw conclusions briefly in Section 5.

\section{Preliminaries}
\subsection{Traveling Salesman Problem (TSP)}
The Traveling Salesman Problem is one of the most intensively studied problems in computational mathematics which is simple to state but very difficult to solve \cite{Applegate:2007:TSP:1374811}. The problem is NP hard, making it not computable in polynomial time \cite{wikiNP}. The  problem  is about finding the shortest possible tour through a set of \texttt{\textit{n}} cities/nodes so  that each city/node is visited exactly once. A weighted graph $G(N,E)$ can represent a TSP, where $N$ represents the cities and $E$ represents the set of edges connecting cities. There is a specific distance $d$ for each $(i,j)\in E$. If $d(i,j) = d(j,i)$, it is known as a symmetric TSP where in an asymmetric TSP, $d(i,j) \ne d(j,i)$ can occur. In our study we consider only the symmetric situation.
\subsection{Ant Colony Systems (ACS)}
Ants, the popular social insects normally live in colonies represent a highly structured distributed system. There are many ant species, some of which are blind. All ant species are known to be deaf \cite{holldobler1990ants}. Despite of these incapabilities, ants are inbred with a strong indirect communication using a chemical produced within them.  While foraging, ants lay the chemical; pheromone on the ground and follow the pheromone placed by other ants. The pheromones tend to decay over time and hence the ants in the colony will choose the path with high pheromone density at the moment. This pheromone communication allows the ants to find the shortest path from the food source to their nest. The optimized behaviors of real ants are based on implementing artificial ant colonies.

The ant colony systems is an example of an ant colony optimization method from the field of swarm intelligence, meta-heuristics and computational intelligence. Around 1990's Marko Dorigo introduced the idea of the ant system and later Dorigo and Gambardella introduce the ant colony system. In the original implementation, ACS was applied to solve the TSP \cite{Dorigo:1997:ACS:2221336.2221407}. Initially, ants in the artificial colony are positioned on random nodes/ cities. They travel from one node to another keeping the travel history in a data structure. The likelihood of selecting a node is based on the pheromone density of the cities laid by other ants, which in the algorithm known as the state transition rule. Once visiting a city, an ant lays an amount of pheromone using the local pheromone updating. Upon completing the tours by all ants, the cities belong to the globally best path again get updated with pheromones using global pheromone updating rule.
\subsubsection{State Transition Rule, Local and Global updating}
State transition rule is responsible for an ant to find its next visiting city. Assume the ant is in the node $r$. It's next city $s$ is determined by the \texttt{equation \ref{eqn:equation 1}}.

 \begin{equation}
 s = \left\{
 \begin{array}{lr}
 arg\quad max_{u\in{J_k(r)}} \{[\tau(r,u)^\theta ]. [\eta(r,u)]^\beta\} &  q\le q_0\\
 S &  Otherwise
 \end{array}
 \right.
 \label{eqn:equation 1}
 \end{equation} \\
where $\tau(r, u)$ is the pheromone density of an edge $(r,u)$, $\eta(r,u)$ is [1/distance$(r, u)$] for TSP. $J_k(r)$ is the set of cities that remain to be visited by ant $k$ positioned on city $r$. The relative importance of the pheromone trail and the heuristic information are represented by the parameters $\theta$ and $\beta$ ($\theta, \beta \ge 0$). $q$ is a random number uniformly distributed in $[0,1]$, $q_0$ is a parameter $(0 \le q_0 \le 1)$, and $S$ is a random variable from the probability distribution given by the \texttt{equation (2)}.

 \begin{equation}
 P_k(r,s) = \left\{
 \begin{array}{lr}
 \dfrac{[\tau(r,u)]^\theta . [\eta(r,u)]^\beta}{\sum_{u\in J_k(r)} [\tau(r,u)]^\theta . [\eta(r,u)]^\beta}  &  if s \in J_k(r)\\
 0 &  Otherwise
 \end{array}
 \right.
 \label{eqn:equation 2}
 \end{equation} \\
ACS local and global updating happens according to the \texttt{equation (3)} and \texttt{equation (4)} respectively.
 \begin{equation}
\tau(r,s) \leftarrow (1-\rho).\tau(r,s) + \rho.\Delta \tau(r,s)
 \label{eqn:equation 3}
 \end{equation} \\
where $0<\rho <1$ is a parameter.\\

 \begin{equation}
 \tau(r,s) \leftarrow (1-\alpha).\tau(r,s) + \alpha.\Delta \tau(r,s)
  \label{eqn:equation 4}
 \end{equation} 
 where  
 \begin{equation}
\Delta \tau(r,s) = \left\{
\begin{array}{lr}
{(L_{gb})}^{-1}  &  if (r,s) \in global \quad best \quad tour\\
0 &  Otherwise
\end{array}
\right.
 \end{equation} \\~\\
$ 0<\alpha<1$ is the pheromone decay parameter and $L_{gb}$ is the length of the globally best tour. In the original implementation, Dorigo et al. have given the set of parameter values obtained from the trial and error approach to suit the selected TSP instances.

The ACS grasped the attention of the world of optimization and hence many researches have been carried out to improve the algorithm as well as to check its ability over solving other optimization problems. 

\subsection{Firefly Algorithm (FA)}
Firefly is a winged beetle commonly known as the lightning bug due to the charming light it emits. The light is used to attract mates or preys. Biological studies reveal many factors about fireflies' life style that are interesting \cite{EVO:EVO1199}. Focusing on their flashing behavior, the firefly algorithm was developed by Xin-She-Yang in 2009 \cite{FA}. The algorithm basically assumes the following.
\begin{itemize}
	\item Fireflies' attraction to each other is gender independent.
	\item Attractiveness is proportional to the brightness of the fireflies, for any two fireflies, the less brighter one is attracted by (and thus moves toward) the brighter one; however, the brightness can decrease as the distance increases; If there is no brighter one than a particular firefly, it moves randomly.
	\item The brightness of a firefly is determined by the value of the problem specific objective function.	
\end{itemize}
As many meta-heuristics, the initial population for the particular problem is generated randomly. In FA also, the parameter set should be specified properly. After these initial steps, the fireflies in the population start moving towards brighter fireflies according to the following equation.
 \begin{equation}
 x_i=x_i+\beta(x_j-x_i )+\alpha(rand-0.5)
 \label{eqn:move}
 \end{equation} \\
 
where
 \begin{equation}
 \beta=\beta_0. e^{-\gamma r^2}
     \label{eqn:attraction}
 \end{equation} \\
 
$\beta_0$ is the attraction at $r=0$. The three terms in \texttt{equation (6)} represent the contribution from the current firefly, attraction between two fireflies and a randomization term respectively. The equation supports both exploitation and exploration. $\alpha$ plays an important role in the randomization process, which is from Uniform or Gaussian distribution. To control the randomness, Yang has used $\delta$, the randomness reduction factor which reduces $\alpha$ according to the \texttt{equation (8)}.
 \begin{equation}
 \alpha=\alpha_0.\delta \quad
 where \quad \delta\in [0,1]
 \label{eqn:reduction}
 \end{equation} \\
FA, as a newcomer in the world of meta-heuristics marked its remarkable capability of handling optimization problems. Yang et al. in 2013 introduced a framework for self-tuning algorithms and it was implemented with the firefly algorithm successfully \cite{yang2013framework}. The framework allows a meta-heuristic algorithm to solve a problem while optimizing it's own algorithm specific parameters.

\section{Self-tuning firefly algorithm optimizing Ant colony system's parameters (ACSFA)}
In this research, the main aim is to tune the parameters of the ant colony system algorithm solving symmetric TSP problems while obtaining the shortest path for the selected TSP. FA as an outstanding performer, has used here in tuning the parameters of the ACS. Another reason is that the self-tuning framework can be easily implemented with firefly algorithm rather than directly on ACS, since the ACS solves a discrete problem (TSP) and the parameters are continuous in nature.

Regarding the ACS, $\beta, \theta, \rho$ and $q_0$ parameters have to be tuned. The FA also has several parameters such as $\alpha, \gamma,\beta$ and the randomness reduction factor $\delta$. The main aim of the self-tuning concept is to find the best parameter settings that minimize the computational cost. When applying the self-tuning framework to the firefly algorithm solving a given optimization problem, both the problem domain and the parameter domain are considered as a single domain in solving the problem. The objective could be the objective of the problem. 

In our case the FA is used to tune the parameters of ACS and the ACS solves a given TSP. The objective of the algorithms is to find the optimal solution of a given TSP instance. The problem of this study contains both parameters of ACS and FA. The pseudo code of the proposed ACSFA algorithm is presented in algorithm \ref{al:algorithm_1}.

After initializing parameters and parameter ranges, inverse of a tour distance is set as the objective of the TSP. The problem is set as a minimization problem. The algorithm will build tours until a predetermined end condition is completed. As in the original ACS, ants are positioned on random starting nodes where each ant completes a tour using the state transition rule stated in \texttt{equation \ref{eqn:equation 1}}. While completing the tour ants will lay pheromones on the visited cities according to the \texttt{equation \ref{eqn:equation 3}}. Upon completing tours by all ants, a globally best tour will be identified and the cities belong to the globally best tour will be awarded with extra pheromone values according to the \texttt{equation \ref{eqn:equation 4}}. Apart from building tours, each ant also carries approximations of the parameters of both ACS and FA.

After completing a tour, all ants will work as fireflies. They now represent approximations of the parameters of FA and ACS. The parameters of ACS and FA will get updated using the self tuning firefly algorithm. The fireflies will move in the direction of better parameters/fireflies according to the \texttt{equation \ref{eqn:move}}. At the end of each tour, the best parameter set will be detected.
\FloatBarrier
\begin{algorithm}[h]
	\caption{: Pseudo code of the ACSFA}
	\label{al:algorithm_1}
	\begin{algorithmic}[1]
		\STATE Begin;
		\STATE Initialize parameter values and ranges of parameter values to be tuned
		\STATE Assign approximations of the parameters to be tuned to each ant (Firefly).
		\STATE Define the objective function ($I - Inverse\quad of\quad the\quad distance$)
		
		\WHILE{End condition}
		
		\STATE Begin Tour
		\STATE \quad Position ants on starting nodes
		\STATE \quad Build the tour while local pheromone update

		\STATE End Tour
		\STATE Do global pheromone update
         \FOR{$ i = 1: n $ (all $n$ ants)}
         \FOR{$ j = 2: n $ ($n$ ants)}
         \IF{$ I_j < I_i $}
         \STATE Move firefly $i$ towards firefly $j$ by using equation (\ref{eqn:move});
         \ENDIF
         \STATE Attractiveness varies with distance $r$ via $e^{-\gamma r^2}$ using equation (\ref{eqn:attraction});
         \STATE Evaluate new solutions and update parameter values;
         \ENDFOR
         \ENDFOR
         \STATE Rank the fireflies and find the current best (best parameter values);
		\ENDWHILE
		\STATE Post process results and visualization;
		\STATE End
	\end{algorithmic}
\end{algorithm}
\FloatBarrier
\section{Experimentation}
The goal of this experiment is to analyze the performance of the ACSFA algorithm solving symmetric TSPs' while selecting the most suitable parameter values for ACS and FA. We aim to formulate the new algorithm with minimum number of user input parameters where all other parameters of the two algorithms are to be tuned while optimizing the TSP instances.
\subsection{Parameter values and problem instances}
The new algorithm is implemented to solve symmetric Traveling Salesman problems. 12 symmetric TSPs are selected for testing \cite{tsplib}. The parameters of ACS include: $\beta, \alpha,\rho, q_0, m, \tau_0$ and $\tau$. $\tau_0$ is based on the nearest neighbor heuristic. Pheromone density $\tau$ depends on $\alpha$ and $\rho$ parameters. Since $\alpha$ works only with the globally best ant, we initialize $\alpha$ to be 0.1. Here, we consider $\beta$, $\rho$ and $q_0$ to be tuned by the self-tuning firefly algorithm. . The range for these parameters are $\beta \in [0 \quad 8]$, $\rho \in [0.5\quad 1]$ and $q_0 \in [0.5\quad 1]$ which are large enough to select the best parameter values for many symmetric TSP instances. Apart from that, we tune the parameters of the FA. The parameters of firefly algorithm include $\alpha, \beta, \gamma, \delta$ and number of fireflies. In equation \ref{eqn:attraction}, the brightness $\beta$ depends on three main factors; $\beta_0$ which we initiate to be 1, $\gamma$; the light absorption coefficient and $r$; the Cartesian distance between two fireflies which should be calculated during iterations. in equation \ref{eqn:reduction}, the value of $\alpha$ get reduced with $\alpha_0$ and $\delta$. $\alpha_0$ is initialized to be 2.3. Therefore the value of $\alpha$ depends on $\delta$. Hence the only factors to be tuned in FA are $\gamma$ and $\delta$. For convenience, we will use $FF\_\alpha_0$, $FF\_\beta_0$, $FF\_\gamma$ and $FF\_\delta$ to indicate the parameters of FA. These parameters varies in the ranges  $FF\_\gamma \in[0\quad10]$ and $FF\_\delta \in[0.8\quad1]$. For further clarification a detailed list of the parameters used for the algorithms are presented in \texttt{Table (\ref{tab: para})}.

\FloatBarrier
\begin{table}[h]
	%\scriptsize
	\centering
	\renewcommand{\arraystretch}{1}
	\setlength\extrarowheight{2pt}
	\begin{tabular}{p{2cm} p{2cm} p{2cm} p{1.5cm} p{2cm} p{1.5cm}}
		\hline
		\multicolumn{2}{l}{\textbf{ACSFA}}  & \multicolumn{2}{l}{\textbf{ACS}} & \multicolumn{2}{l}{\textbf{PSOACS}}  \\ \hline
		Parameter & \parbox{1.5cm}{Value/\\Range}           & Parameter    & \parbox{1.5cm}{Value/\\Range}  & Parameter      & \parbox{1.5cm}{Value/\\Range}      \\ \hline
		$\alpha$ & 0.1             & $\alpha$ & 0.1     & 	$\beta$ & $[0\quad8]$     \\ 
		$\beta$  & $[0\quad8]$             &    $\beta$  & 2  & 	$\rho$ & $[0.5\quad1]$           \\
		$\rho$ & $[0.5\quad1]$   &  $\rho$  &  0.1       &  $q_0$ & $[0.5\quad1]$             \\
		$q_0$   & $[0.5\quad1]$ &    &         &    $PSO\_Q_1$           &   2            \\
		$FF\_\alpha_0$  & 2.3 &     &        &   $PSO\_Q_2$                 &   2             \\ 
		$FF\_\beta_0$   & 1 &    &          &                &                  \\
		$FF\_\gamma$   & $[0\quad10]$ &   &         &                &                  \\
		$FF\_\delta$   & $[0.8\quad1]$ &   &         &                &                  \\\hline
	\end{tabular}
	\caption{Parameter values and ranges used for ACSFA, ACS and PSOACS}
	\label{tab: para}
\end{table}
\FloatBarrier

 The new algorithm is implemented using MATLAB \cite{MATLAB:2010} and the experiment is conducted on a laptop with an Intel (R) Core (TM) i5-5200U CPU @ 2.20 GHz processor and 8GB memory. Since the speed relies on the programming language, structure and the type of the machine, the comparing algorithms were also implemented using the same environment. The new algorithm is compared with the original ant colony algorithm (ACS) \cite{Dorigo:1997:ACS:2221336.2221407} and an adaptive parameter control strategy for ACO implemented using the PSO algorithm (PSOACS) \cite{4028059}. The TSP instances and their best known solutions are indicated in \texttt{Table (\ref{Tab:2})}.

\FloatBarrier
\begin{table}[h]
	\centering
		\renewcommand{\arraystretch}{1}
		\setlength\extrarowheight{2pt}
	\begin{tabular}{ll}
		\hline
		\textbf{TSP Instance} & \textbf{Optimal Tour Length}  \\ \hline
		ulysses16 & \hspace*{1cm}6859 \\
		bays29 & \hspace*{1cm}2020  \\
		Oliver30 & \hspace*{1cm}420  \\
		eil51 & \hspace*{1cm}426  \\
		pr76 & \hspace*{1cm}108159  \\
		kroA100 & \hspace*{1cm}21282  \\
		lin105 & \hspace*{1cm}14379 \\
		tsp225 & \hspace*{1cm}3916 \\
		gil262 & \hspace*{1cm}2378 \\
		lin318 & \hspace*{1cm}42029 \\
		rat575 & \hspace*{1cm}6773  \\
		rat783 & \hspace*{1cm}8806 \\ \hline
	\end{tabular}
		\caption{Problem instances and the Optimal tour lengths found so far}
		\label{Tab:2}
\end{table}
\FloatBarrier
\section{Results}
To accomplish the experimental comparison, we considered randomly selected 12 TSP instances from the TSPLIB, that have been presented in the \texttt{Table (\ref{Tab:2})} \cite{tsplib}. \texttt{Table (\ref{Tab:3})} presents the results. Each algorithm executed 10 times with each TSP instance to get the results. Results are formatted as the best TSP distance, the average, the worst and the average time taken by each algorithm. The obtained results illustrate the strength of the ACSFA over other two algorithms. The interesting factor is that, in ACSFA, results are better and most of the parameters are handled by the self tuning firefly algorithm.
\FloatBarrier
\begin{table}[h]
	\centering
	\footnotesize
		\renewcommand{\arraystretch}{1}
		\setlength\extrarowheight{1.7pt}
	\begin{tabular}{|p{2.5cm}|p{2cm}|p{1.5cm}p{1.5cm}p{1.5cm}p{1.5cm}|}
		\hline
		TSP Instance& Algorithm & Best & Average & Worst & $t_{avg}$ (s) \\ \hline \hline
		\multirow{3}{*}{ulysses16} & ACSFA & 6859 & 6891.2 & 6909 & 11.02 \\ \cline{2-6} 
		& PSOACS & 6909 & 6909 & 6909  & 11.47  \\ \cline{2-6} 
		& ACS & 6875 & 6891.2 & 6909 & 11.02 \\ \hline \hline
		\multirow{3}{*}{bays29} & ACSFA & 2026 & 2065 & 2085 &36.58  \\ \cline{2-6} 
		& PSOACS & 2028 &2033.6  & 2036 & 36.87 \\ \cline{2-6} 
		& ACS & 2038 & 2041.25 & 2042 & 33.30 \\ \hline \hline
		\multirow{3}{*}{Oliver30} & ACSFA & 421 & 425 & 426 & 22.27 \\ \cline{2-6} 
		& PSOACS & 425 & 425.5 & 426 & 23.85 \\ \cline{2-6} 
		& ACS & 426 & 428.83 &434  &19.77  \\ \hline \hline
		\multirow{3}{*}{eil51} & ACSFA & 428  &432.6  & 438 & 67.30 \\ \cline{2-6} 
		& PSOACS & 429 & 429.8 & 431 & 70.40  \\ \cline{2-6} 
		& ACS &  430& 434 & 439 & 60.16 \\ \hline \hline
		
			\multirow{3}{*}{pr76} & ACSFA & 108358 & 108474 & 108644 & 116.80 \\ \cline{2-6} 
			& PSOACS & 108358 & 103755.8 & 110488 & 157.92  \\ \cline{2-6} 
			& ACS & 110281 & 111342.6 & 112714 & 91.16 \\ \hline \hline
			\multirow{3}{*}{kroA100} & ACSFA &  21396 &  21390.71& 21537 & 163.67 \\ \cline{2-6} 
			& PSOACS & 21835 & 21874 & 21914 & 180.48  \\ \cline{2-6} 
			& ACS & 22011  & 22703.16 & 23385 & 131.25 \\ \hline \hline
			
		\multirow{3}{*}{lin105} & ACSFA & 14412  &14706.25  &  14860 & 211.09  \\ \cline{2-6} 
		& PSOACS & 14492 & 14503.33 & 14571  & 194.43 \\ \cline{2-6} 
		& ACS &  14844 & 15051& 15353 & 153.10\\ \hline \hline
		
		\multirow{3}{*}{TSP225} & ACSFA & 3978   &4107.8 &  4283 & 519.57  \\ \cline{2-6} 
		& PSOACS & 4009 & 4039.25 & 4059 &  611.89\\ \cline{2-6} 
		& ACS & 4077  &4220.2 & 4308 & 395.20\\ \hline \hline
		
		\multirow{3}{*}{gil262} & ACSFA &2435   & 2448.16& 2471  &  686.22 \\ \cline{2-6} 
		& PSOACS &2442  & 2472 & 2488 &788.59  \\ \cline{2-6} 
		& ACS & 2722  &2872 & 2859 & 489.55\\ \hline \hline
		
		\multirow{3}{*}{lin318} & ACSFA & 43061  &43718.6 & 44192  & 1031.35  \\ \cline{2-6} 
		& PSOACS &43191  & 43737.6 &44211  & 1156.92 \\ \cline{2-6} 
		& ACS & 47960  & 48445 & 49427 &595.66 \\ \hline \hline

		\multirow{3}{*}{rat575} & ACSFA &  7097 & 7420.33&7674   & 2952.31  \\ \cline{2-6} 
		& PSOACS & 7189  &7242.33  & 7346 & 3373.18 \\ \cline{2-6} 
		& ACS & 7819  &7877.5 &7965  &1609.39 \\ \hline \hline

		\multirow{3}{*}{rat783} & ACSFA & 10067  & 10226.2&  10382 & 5998.23  \\ \cline{2-6} 
		& PSOACS & 10540 & 10540 & 10540 & 6216.225 \\ \cline{2-6} 
		& ACS & 10165  &10491.25 & 10642 & 5143.26 \\ \hline 		
		
	\end{tabular}
		\caption{The best, average , worst performance and the average time of the algorithms}
		\label{Tab:3}
\end{table}
\FloatBarrier
 However, to prove the results in a convenient way, a statistical analysis is also conducted over the obtained results. 
\subsection{Statistical Analysis}
A convenient statistical analysis was conducted to prove the validity of the results.The guidelines provided by Derrac et al. \cite{Derrac2013} were followed to perform the statistical analysis. However, here we have used parametric methods to conduct the statistical comparison. To test the hypothesis' related to the study, we have used ANOVA (Analysis of variance) technique \cite{winer}. ANOVA is useful when we need to do an experiment to conduct a comparison over more than 2 samples. Therefore, to compare the 3 Algorithms in terms of \textbf{error}, ANOVA with RCBD (Randomized Complete Block Designs) has been applied. The hypothesis tested here is:\\
\texttt{$H_0$: There is no any significant difference between 3 algorithms}\\
\texttt{$H_1$: At least one algorithm is different from others}

\noindent The results of the performed statistical test is as follows:\\

\noindent\textbf{\underline{Analysis for error}}\\~\\
\small
\texttt{Method}\\
\texttt{Factor coding  (-1, 0, +1)}\\
\texttt{Factor Information}\\~\\
\begin{tabular}{p{2cm}p{2cm}p{2cm}p{8cm}}
	\texttt{Factor} & \texttt{Type} & \texttt{Levels} & \texttt{Values}\\
	\texttt{Algorithm} & \texttt{Fixed} & \texttt{ 3} & \texttt{ACS, ACSFA, PSOACS}\\
	\texttt{TSP} & \texttt{Fixed} & \texttt{ 12} & \texttt{Bays29, eil51, gil262, kroA100, lin105, lin318, Oliver30, pr76,
		rat575, rat783, TSP225, ulysses16}\\
\end{tabular}
\noindent
\texttt{Analysis of Variance}\\
\noindent
\begin{tabular}{p{2cm}p{2cm}p{2cm}p{2cm}p{2cm}p{2cm}}
\texttt{Source} & \texttt{DF} & \texttt{Adj SS} & \texttt{Adj MS} & \texttt{F - Value} & \texttt{P - Value}\\
\texttt{Algorithm}  & \texttt{2} &  \texttt{4043366} & \texttt{2021683}  &  \texttt{3.00}  &  \texttt{0.070} \\
\texttt{TSP} & \texttt{11}  &\texttt{21614129} & \texttt{1964921} & \texttt{2.92}  &\texttt{0.016}\\
\texttt{Error} & \texttt{22} & \texttt{14808697} & \texttt{673123}\\
\texttt{Total} & \texttt{35} & \texttt{40466192}\\~\\
\end{tabular}	    
\noindent
\texttt{Model Summary}\\
\begin{tabular}{p{2cm}p{2cm}p{2cm}p{2cm}}
	\texttt{S} & \texttt{R-sq} & \texttt{R-sq(adj)} & \texttt{R-sq(pred)} \\
	\texttt{820.440}  &	\texttt{63.40\%}  & \texttt{41.78\%}  & \texttt{2.01\%}	\\~\\
\end{tabular}

	Table 5:  Results of ANOVA for the `error'\\

\FloatBarrier

P value of the algorithm field (0.07) is less than 0.1. Based on that, we reject $H_0$  and the conclusion can be made as that there exist at least one algorithm which is significantly different from others at a 0.1 significance level. To find which algorithms are different from which, we have conducted a pairwise comparison over the algorithms. To conduct a pairwise comparison over the error, Tukey's method was applied \cite{wikiTukey}. The results were as follows.\\~\\ ~\\
\noindent\textbf{\underline{Tukey Pairwise Comparisons: Response = error, Term = Algorithm }}\\~\\
\small
\texttt{Grouping Information Using the Tukey Method and 90\% Confidence}
\FloatBarrier
\begin{table}[h]\small
\begin{tabular}{p{2cm}p{2cm}p{2cm}p{2cm}}
	\texttt{Algorithm} & \texttt{N} & \texttt{Mean} & \texttt{Grouping} \\
    \texttt{ACS}   & 	\texttt{12}  &	\texttt{1016.75}  &	\texttt{A}\\
    \texttt{PSOACS} &  	\texttt{12} &  	\texttt{366.67} &  	\texttt{A      B}\\
    \texttt{ACSFA}  & \texttt{12} &  \texttt{257.58}  &   \texttt{B}\\	
\end{tabular}
\end{table}
\FloatBarrier \noindent
\small \texttt{Means that do not share a letter are significantly different.}\\
\FloatBarrier
\begin{figure}[h]
	\includegraphics[scale=0.72]{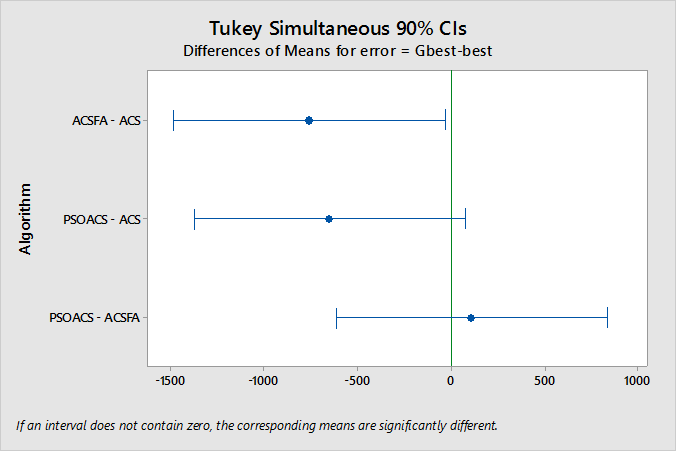}
	\caption{Tukey Pairwise Comparisons for `error'}
\end{figure}
\FloatBarrier

 The hypothesis used this time is as follows.\\~\\
 \texttt{$H_0$: There is no any difference between 2 algorithms}\\
 \texttt{$H_1$: There is a difference between 2 algorithms}\\

The pairwise comparison on the error concluded that there is no any difference between ACSFA and PSOACS at 90\% confidence level and there is a difference between ACSFA and ACS at 90\% confidence level.

The same procedure was conducted considering the \textbf{average} and the \textbf{best} results of the algorithms for 12 TSP instances. Having P- values as 0.072 and 0.070, Analysis of variance in both cases supported to reject $H_0$ concluding that  at least one algorithm is significantly different from others at a 0.1 significance level. Pairwise comparisons were also conducted in both cases. The results are as follows.\\

\noindent\textbf{\underline{Tukey Pairwise Comparisons: Response = Average, Term = Algorithm }}\\~\\
\small
\texttt{Grouping Information Using the Tukey Method and 90\% Confidence}
\FloatBarrier
\begin{table}[h]\small
	\begin{tabular}{p{2cm}p{2cm}p{2cm}p{2cm}}
		\texttt{Algorithm} & \texttt{N} & \texttt{Mean} & \texttt{Grouping} \\
		\texttt{ACS}   & 	\texttt{12}  &	\texttt{19399.8}  &	\texttt{A}\\
		\texttt{PSOACS} &  	\texttt{12} &  	\texttt{18525.5} &  	\texttt{A      B}\\
		\texttt{ACSFA}  & \texttt{12} &  \texttt{18163.5}  &   \texttt{B}\\	
	\end{tabular}
\end{table}
\FloatBarrier \noindent
\small \texttt{Means that do not share a letter are significantly different.}\\
\FloatBarrier
\begin{figure}[h]
	\includegraphics[scale=0.72]{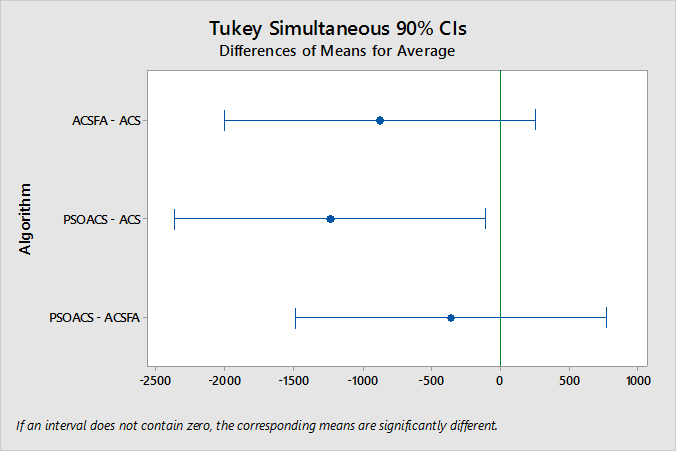}
	\caption{Tukey Pairwise Comparisons for `average'\\~\\}
\end{figure}
\FloatBarrier

\noindent\textbf{\underline{Tukey Pairwise Comparisons: Response = Best, Term = Algorithm}}\\~\\
\small
\texttt{Grouping Information Using the Tukey Method and 90\% Confidence}
\FloatBarrier
\begin{table}[h]\small
	\begin{tabular}{p{2cm}p{2cm}p{2cm}p{2cm}}
		\texttt{Algorithm} & \texttt{N} & \texttt{Mean} & \texttt{Grouping} \\
		\texttt{ACS}   & 	\texttt{12}  &	\texttt{19137.3}  &	\texttt{A}\\
		\texttt{PSOACS} &  	\texttt{12} &  	\texttt{18487.2} &  	\texttt{A      B}\\
		\texttt{ACSFA}  & \texttt{12} &  \texttt{18378.2}  &   \texttt{B}\\	
	\end{tabular}
\end{table}
\FloatBarrier \noindent
\small \texttt{Means that do not share a letter are significantly different.}
\FloatBarrier
\begin{figure}[h]
\includegraphics[scale=0.72]{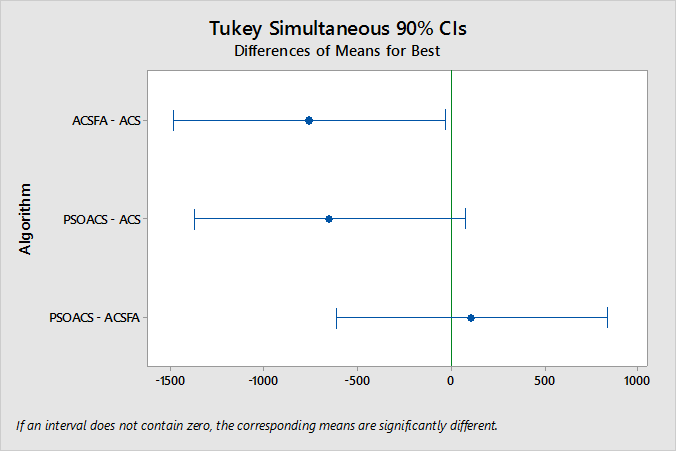}\\
\caption{Tukey Pairwise Comparisons for `best results'\\~\\}
\end{figure}
\FloatBarrier
Both states support the conclusion that there is no any difference between ACSFA and PSOACS at 90\% confidence level. Considering the best case, ACSFA and ACS appeared to be different at 90\% confidence level.

Finally the analysis support the facts that the performance of ACSFA and PSOACS is equally strong where the performance of ACS is not as strong as ACSFA and PSOACS. But the results considering the best case, demonstrate that ACSFA outperforms other two algorithms. However although there is no significant difference between ACSFA and PSOACS from the statistical viewpoint, there exists a strong advantage of ACSFA over PSOACS: the ability of performing well without considering the selection of suitable parameter values for both ACS and FA.
\subsection{Parameter Optimization}
\begin{figure}[h]
	\centering
	\includegraphics[scale=0.8]{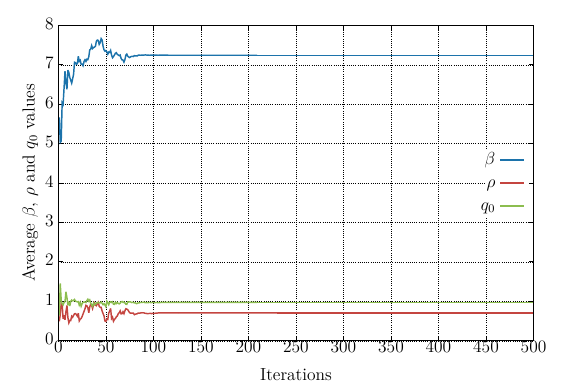}
	
	\caption{Variation of the average $\beta$,$\rho$ and $q_0$ values over iterations for eil51 TSP instance}
	\label{fig: avgBetaRawQ0}
\end{figure}

The statistical study emphasizes that there is no significant difference between ACSFA and PSOACS. But still ACSFA is better since it does provide a parameter-free environment to the user. The firefly algorithm with the self-tuning framework tunes the necessary parameters of ACS as well as FA. The figure \ref{fig: avgBetaRawQ0} represents the evolution of parameter values of ACS over the iterations for the eil51 TSP instance. Here the mean value of each parameter for each iteration is calculated.

\FloatBarrier
\begin{figure}[h]
	\centering
	\includegraphics[scale=0.8]{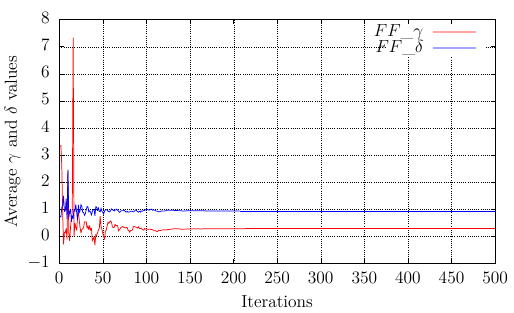}
	
	\caption{Variation of the average $FF\_\gamma$ and $FF\_\delta$ values over iterations for eil51 TSP instance}
	\label{fig: avgGammaDelta}
\end{figure}

\FloatBarrier
It is necessary to see the behavior of the parameters of the FA as well. Figure \ref{fig: avgGammaDelta} shows the evolution of the parameters of FA during iterations for the eil51 TSP instance. Here also, the mean value of each parameter for each iteration is calculated. 

\FloatBarrier
\FloatBarrier

Figures \ref{fig: avgBetaRawQ0} and \ref{fig: avgGammaDelta} point out that the parameter values varies up to some number of iterations and then stabilize over an optimum value.  
\section{Concluding Remarks}
The study has focused on implementing an ant colony algorithm to solve symmetric TSP problems whose parameters are handled by a self tuning firefly algorithm. The algorithm was successfully implemented and tested with standard TSP problems. According to the results obtained, some key conclusions can be drawn. In terms of optimization, the results show that the ACSFA performs well in finding the shortest path for a given TSP instance. The comparisons done with ACS and PSOACS shows ACSFA works well. Although the statistical analysis concludes that both ACSFA and PSOACS have same performance, ACSFA outperforms PSOACS by providing a parameter free environment. The self tuning framework worked fine with the firefly algorithm in tuning both parameters of ACS and FA. The graphical representations of the evolution of parameters of both ACS and FA clearly demonstrates the ability of the self tuning firefly algorithm. With these we can consider the new ACSFA as a better performer to solve TSPs using ACS. \\
For further development, this research encourages us to study the performance of the self tuning framework with other nature inspired algorithms such as particle swarm optimization, bees algorithm etc. Also since the increasing number of cities drops the performance of the algorithm, more experimentation should be done on the population size and the initialization of parameter ranges as well.   
\section*{Acknowledgment}
	The authors would like to thank Dr. Xin-She Yang for his
	valuable suggestions and explanations on implementing the
	self tuning framework and Ms. W.J. Polegoda, lecturer at
	the Faculty of Animal Science \& Export Agriculture, Uva
	Wellassa University, Sri Lanka for the guidance given on
	the statistical analysis.
	\vspace*{-1cm}
\section*{}

\end{document}